\documentclass{arxivtmpl}

\def\1{\bm{1}}

\DeclareMathAlphabet{\mathsfit}{\encodingdefault}{\sfdefault}{m}{sl}
\SetMathAlphabet{\mathsfit}{bold}{\encodingdefault}{\sfdefault}{bx}{n}

\hypersetup{
    colorlinks=true,
    linkcolor=linkcol,
    citecolor=linkcol,
    urlcolor=linkcol,
    pdfborderstyle={},
    pdfborder={0 0 0}
}
\renewcommand\Affilfont{\rmfamily\mdseries\fontsize{11.5}{14}\selectfont}

\colorlet{perfhue}{gblue}
\colorlet{perf0}{white}
\colorlet{perf10}{perfhue!12}
\colorlet{perf20}{perfhue!22}
\colorlet{perf30}{perfhue!33}
\colorlet{perf40}{perfhue!44}
\colorlet{perf50}{perfhue!55}
\colorlet{perf60}{perfhue!66}
\colorlet{perf70}{perfhue!77}
\colorlet{perf80}{perfhue!88}

\newcommand{\RN}[1]{%
	\textup{\lowercase\expandafter{\it \romannumeral#1}}%
}

\SetKwInput{KwInput}{Input}
\SetKwInput{KwOutput}{Output}
\DontPrintSemicolon

\SetKwComment{Comment}{\color{ggreen}\# }{}

\SetKwProg{Function}{def}{:}{}

\SetKwProg{For}{for}{:}{}
\SetKwProg{If}{if}{:}{}

\usepackage{url}

\definecolor{spblue}{HTML}{123F8C}
\definecolor{spsteel}{HTML}{1F6FD6}
\definecolor{spcoral}{HTML}{E8613C}
\definecolor{sprow}{gray}{0.9}
\definecolor{spgray}{HTML}{9AA1AB}
\hypersetup{linkcolor=spblue,citecolor=spblue,urlcolor=spblue}

\newcommand{\kj}[1]{}
\newcommand{\runin}[1]{\textbf{#1}\hspace{0.5em}}

\newcommand{\panelrule}[1]{\hfill\raisebox{4pt}{\textcolor[HTML]{DADCE0}{\rule{0.6pt}{#1}}}\hfill}
\title{SpecPilot: Autonomous Algorithm Discovery and Systems Optimization for Speculative Decoding}

\author[1,$*$]{Kunjun Li}
\author[2,$*$]{Wenhao Chai}
\author[2,$*$]{Yihao Liang}
\author[1]{Yuezhou Hu}
\author[3]{Shuyao Xu}
\author[1]{Yi Pan}
\author[1]{William X. Zheng}
\author[1]{Kannan Ramchandran}
\author[1]{Kurt Keutzer}
\author[1,4]{Jiantao Jiao}
\affil[1]{University of California, Berkeley}
\affil[2]{Princeton University}
\affil[3]{Stanford University}
\affil[4]{NVIDIA}
\renewcommand{\institutionlogos}{%
  \raisebox{-0.5\height}{\includegraphics[height=24pt]{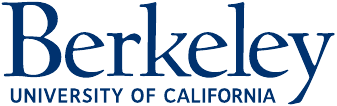}}\hspace{16pt}%
  \raisebox{-0.5\height}{\includegraphics[height=22pt]{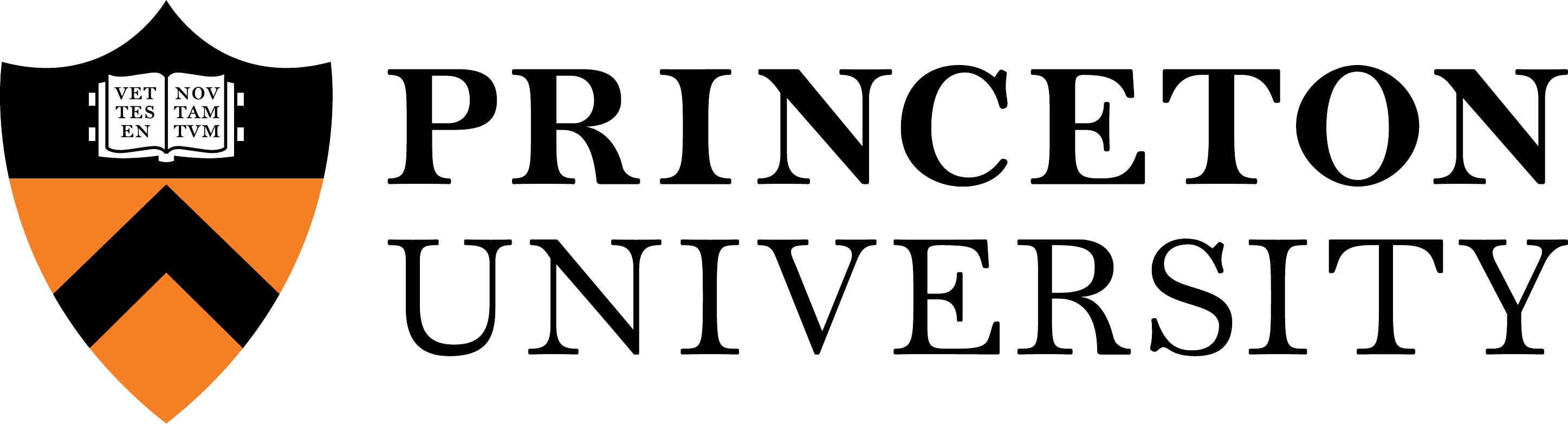}}\hspace{16pt}%
  \raisebox{-0.5\height}{\includegraphics[height=24pt]{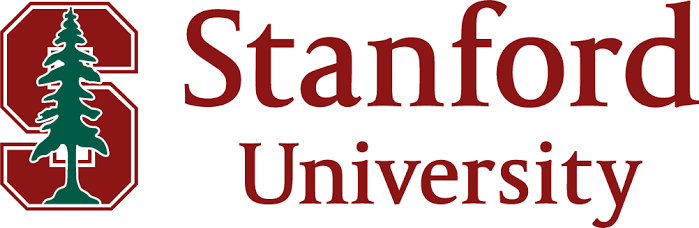}}\hspace{16pt}%
  \raisebox{-0.5\height}{\includegraphics[height=18pt]{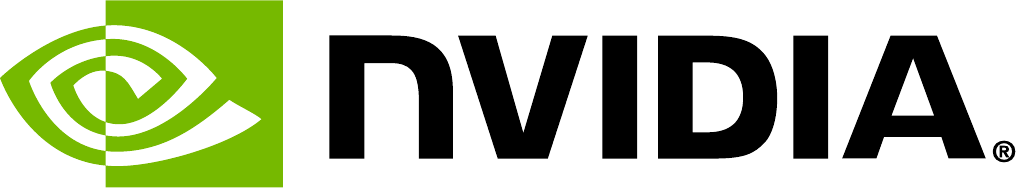}}}
\fancypagestyle{firstpagestyle}{\fancyhf{}\fancyhead[L]{\institutionlogos\rule[-18pt]{0pt}{0pt}}%
  }
\renewcommand{\titlefont}{\normalfont\bfseries\fontsize{17}{21}\selectfont}
\renewcommand\Affilfont{\rmfamily\mdseries\fontsize{10}{12.5}\selectfont}
\authornote{$*$\,Equal contribution.\quad Corresponding Author: \href{mailto:jiantaoj@nvidia.com}{\texttt{jiantaoj@nvidia.com}}\\[5pt]
Website: \href{https://stargazerx0.github.io/specpilot-project-page/}{\texttt{Specpilot-Project-Page}}\quad
Code: \href{https://github.com/StargazerX0/SpecPilot}{\texttt{SpecPilot}}}

\begin{document}

\maketitle

\thispagestyle{firstpagestyle}

\begin{abstract}
{\rmfamily\bfseries\normalsize Abstract}\par\vspace{0.15\baselineskip}
\small
Speculative decoding accelerates memory-bound LLM decoding with a lightweight drafter, and parallel drafters extend this speedup by drafting a whole block in one forward pass. However, it remains unclear whether their released systems (\emph{stock}) attain the full speedup that parallel drafting permits. We identify three challenges. First, with one request, each step accepts few tokens but reads every target weight. Second, with batching, per-request verification work roughly halves the speedup over autoregressive (AR) decoding. Third, speculative decoding is deployment-specific, and no universal optimization fits every serving regime, drafter, target model and GPU. To address these challenges, we present SpecPilot, an autonomous agentic loop that discovers algorithms and systems optimizations for each deployment. In each round, a cost model locates the largest gaps between the current system and its ceiling. Parallel agents explore directions and integrate the successful ones. Token forcing checks every output token against stock to detect reward hacking. A structured memory carries validated mechanisms to new rounds and deployments. SpecPilot discovered a general lossless 12-bit weight format, tree verification for hybrid recurrent targets, two-level verification and three batching mechanisms. We evaluate it across four target models, three drafters, two GPU platforms and five benchmarks. Without training or quality loss, it raises the speedup over AR from stock's 4.46$\times$ to 10.87$\times$ with one request, and from 1.94$\times$ to 3.93$\times$ with batching.\looseness=-1\par
\end{abstract}
\vspace{0.75em}

\setcounter{topnumber}{1}
\section{Introduction}

\label{sec:intro}
\suppressfloats[t]

Speculative decoding~\cite{leviathan2023fast,chen2023accelerating} accelerates memory-bound autoregressive (AR) decoding of large language models (LLMs) with a lightweight drafter, so one target pass accepts several tokens. AR drafters such as EAGLE-3~\cite{li2025eagle3} draft sequentially, while parallel drafters such as DFlash~\cite{dflash2026}, DFlash2~\cite{inco2026dflash2} and DSpark~\cite{dspark2026} draft a block of 8 or 16 tokens per forward pass.

\begin{figure}[t]

\centering

\includegraphics[width=\textwidth]{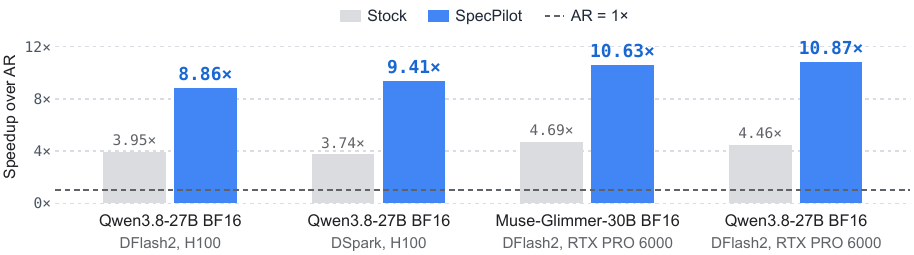}

\vskip -0.08in

\caption{Speedup over AR with one request. SpecPilot more than doubles stock's speedup in each setting.}

\label{fig:headline}

\vskip -0.15in

\end{figure}

\begin{figure*}[t]

\centering

\includegraphics[width=\textwidth]{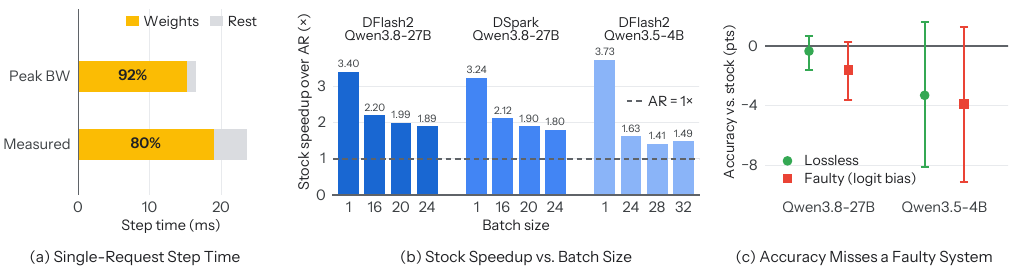}

\vskip -0.08in

\caption{(a) Single-request step time split into target weights and the rest, at peak bandwidth and as measured after the first optimization round. Reading the weights dominates the step. (b) Stock speedup over AR falls as the batch size grows, for Qwen3.8-27B in FP8 and Qwen3.5-4B in BF16. (c) Task accuracy change against stock, with 95\% CIs, for a lossless system and a faulty one with an injected logit bias. All four intervals include zero, so accuracy cannot flag the faulty systems.}

\label{fig:motiv}

\vskip -0.1in

\end{figure*}

However, their released systems (\emph{stock}) leave much of this potential unrealized. With one request, reading the target weights dominates the per-step latency (Figure~\ref{fig:motiv}a), so verifying additional tokens adds little extra latency. Stock still verifies only one drafted chain, ignoring the other candidates that its drafter ranks at each position, which often include the target's next token. Using them would accept about twice as many tokens per step (\S\ref{sec:model}). With batching, each request adds its own attention, recurrent-state and prefill work, so weight reads no longer dominate the step, and stock's speedup over AR decoding roughly halves (Figure~\ref{fig:motiv}b).

Existing techniques such as draft trees~\cite{miao2024specinfer,chen2024sequoia,li2024eagle2,ringel2026ddtree} and batch-aware speculation~\cite{liu2024goodput,chen2024magicdec,dcut2026} are general, but the best choice depends on the deployment. A \emph{deployment} comprises a serving regime, a drafter, a target and a GPU platform, and each changes what pays off. The serving regime can turn a gain into a loss. Checking drafts with a low-precision target copy before the exact pass speeds up decoding with one request, but with batching on BF16 Qwen3.8-27B the same idea is 13\% slower, because the extra check takes nearly a step. The drafter decides how to score a draft tree. DSpark conditions each drafted token on the one before it, and a tree that ignores this accepts half the tokens of one using it. The target and the GPU decide what each mechanism costs. A tree on hybrid Qwen3.8-27B~\cite{qwen2026qwen38} must track every node's recurrent state (\S\ref{sec:bg-hybrid}), and the RTX PRO 6000 lacks Hopper's \texttt{wgmma} instructions, so kernels need porting. No universal optimization fits every deployment.

\suppressfloats[t]

Manually optimizing each deployment does not scale, whereas coding agents~\cite{anthropic2025claudecode,openai2025codex} iterate rapidly. This makes per-deployment optimization a natural fit for them, and agentic optimizers such as AlphaEvolve and ADRS already evolve programs against automated evaluators~\cite{novikov2025alphaevolve,cheng2025adrs}. For lossless speculative decoding, however, building a reliable evaluator is difficult for two reasons. First, many gains are only a few percent, close to timing noise, and a gain measured on development requests may not generalize to unseen ones. Second, an agent rewarded for speed is prone to reward hacking~\cite{amodei2016concrete,skalse2022defining}, such as adopting a faster numerical shortcut that changes outputs without a measurable drop in task accuracy (Figure~\ref{fig:motiv}c).

\begin{figure*}[t]

\centering

\includegraphics[width=\textwidth]{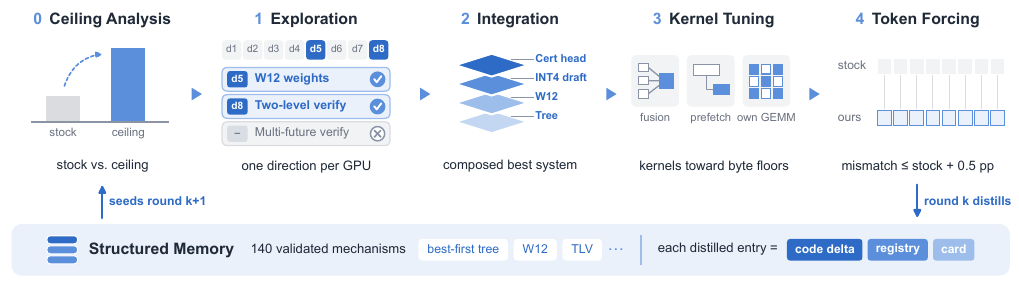}

\vskip -0.05in

\caption{One round of the SpecPilot loop. \textbf{0}~The cost model measures the gap between the current system and its ceiling. \textbf{1}~Directions run in parallel, one per GPU. \textbf{2}~Validated pieces stack into one system. \textbf{3}~A kernel sub-agent tunes the kernels furthest from their byte floors. \textbf{4}~Token forcing checks the sealed final against stock at every teacher-forced token. Each round distills its mechanisms into memory entries, which carry to later rounds and new deployments.}

\label{fig:agent}

\vskip -0.1in

\end{figure*}

We present SpecPilot (Figure~\ref{fig:agent}), an autonomous agentic loop that searches for speculative decoding optimizations tailored to each deployment. It rests on four ideas. First, a cost model compares measured performance with an estimated ceiling to locate where speed is lost. Second, parallel agents explore multiple directions on separate GPUs and integrate the validated pieces, so small gains add up. Third, a controller isolated from the agents accepts a change only on a significant gain confirmed on unseen requests, and token forcing checks every output token against stock to detect lossy changes (\S\ref{sec:forcing}). Finally, validated optimizations are distilled into a structured memory as replayable code and evidence, for reuse in later rounds and new deployments.\looseness=-1


The loop developed lossless techniques, each targeting a gap the cost model measured (\S\ref{sec:built}). Best-first tree verification exploits idle drafter candidates, even on hybrid recurrent targets, and W12, a general lossless 12-bit weight format, cuts the weight reads dominating single-request steps. Two-level verification pre-checks drafts at low precision, nearly doubling accepted tokens per exact pass. For batching, three mechanisms spend verify rows where they pay off (\S\ref{sec:load}).

We evaluate SpecPilot on four target models, three drafters, two GPU platforms and five benchmarks. Without training or quality loss, it raises the speedup over AR from stock's 4.46$\times$ to 10.87$\times$ with one request (Figure~\ref{fig:headline}), and from 1.94$\times$ to 3.93$\times$ with batching. On H100 with one request, it is 1.84$\times$ faster than DDTree~\cite{ringel2026ddtree}.

\newpage
In summary, this paper makes the following contributions:

\begin{itemize}[topsep=2pt,partopsep=0pt,itemsep=1pt,parsep=0pt]

\item Autonomous Optimization Loop: We design SpecPilot, a ceiling-guided agentic framework that tailors speculative decoding to each deployment and transfers validated optimizations across hardware and models via a structured memory. To prevent reward hacking, we introduce token forcing, a stock-calibrated test that detects output divergences task accuracy misses.

\item Discovered Lossless Algorithms: SpecPilot discovered W12, a general lossless 12-bit weight format, best-first tree verification for hybrid recurrent targets, two-level verification and three mechanisms for batching.

\item Cross-Regime Bottleneck Analysis: We show how ceiling-guided discovery adapts to both memory-bandwidth and per-row compute bottlenecks, and why the same mechanism can accelerate one request but slow decoding with batching on the same GPU.

\end{itemize}

\section{Background}

\label{sec:background}

\subsection{Speculative Decoding}

\label{sec:bg-sd}

In speculative decoding~\cite{leviathan2023fast,chen2023accelerating}, a drafter proposes $m$ tokens $\hat{x}_{1..m}$ after the prefix $x_{<t}$, and the target model verifies them in one forward pass, one \emph{row} per token. Greedy decoding accepts the longest prefix with $\hat{x}_i=\arg\max p(\cdot\mid x_{<t},\hat{x}_{<i})$ and adds the target's argmax as a bonus, accepting $\tau\in[1,m+1]$ tokens.

\runin{Lossless.}A system is lossless if every accepted token lies on a path verified by a pass over the exact target weights (\emph{exact pass}) and equals that pass's argmax. Its output is then the target's greedy decode up to floating-point reduction order, such as GEMM tiling or tree attention, which can change a token only at a logit near-tie. Stock already differs from AR in this way. At each first divergence, the two tokens' log-probabilities differ by at most 0.232 nats (\S\ref{sec:forcing}). A low-precision verify pass, by contrast, can change tokens even far from near-ties. This makes losslessness hard to test. Requiring identical outputs rejects lossless systems, and task accuracy misses faulty ones (Figure~\ref{fig:motiv}c). Token forcing (\S\ref{sec:forcing}) instead checks every token statistically against stock.

\subsection{Parallel Drafters and Draft Trees}

\label{sec:bg-parallel}

\label{sec:bg-trees}

DFlash2~\cite{inco2026dflash2}, like DFlash~\cite{dflash2026}, drafts a block of masked positions in one pass from target hidden states. A low-rank selector scores each of the top-16 candidates per position given one at the previous position, and candidates and scores form the \emph{lattice}. DSpark~\cite{dspark2026} adds a Markov head on the previous token:
\begin{equation}
\ell_s(y) = \mathrm{base}_s(y) + \big\langle W_1[x_{s-1}],\, W_2[y]\big\rangle,
\label{eq:markov}
\end{equation}
where $\mathrm{base}_s$ denotes the backbone's logits, and stock DSpark drafts a greedy chain with it.

A draft tree contains continuations that the target verifies in one pass, each node seeing only its ancestors, and the step accepts the longest root path matching the target's argmax plus a bonus token~\cite{miao2024specinfer}. DDTree~\cite{ringel2026ddtree} adds nodes by path log-probability up to $N$, so every prefix of this order is a best-first tree (\emph{prefix-closed}).

\subsection{Hybrid Targets}

\label{sec:bg-hybrid}

Our main target model, Qwen3.8-27B~\cite{qwen2026qwen38}, uses Gated DeltaNet (GDN)~\cite{yang2025gated} in 48 of 64 layers and attention elsewhere. A GDN head updates a fixed-size state $h$ per token $t$ by the gated delta rule,
\begin{equation}
h_t = \alpha_t\big(I-\beta_t k_t k_t^{\top}\big)h_{t-1} + \beta_t k_t v_t^{\top}, \qquad o_t = h_t^{\top} q_t,
\label{eq:gdn}
\end{equation}
with decay gate $\alpha_t$, write strength $\beta_t$, and key, value and query $k_t, v_t, q_t$. Stock verification snapshots $h$ at each verified position, so a tree needs one snapshot per node.

\section{SpecPilot}

\label{sec:loop}

\label{sec:method}

SpecPilot operates in \emph{rounds} (Figure~\ref{fig:agent}). A round measures where the current system loses speed (Phase~0), lets agents build and time changes (Phases~1--3) and ends with a sealed final on unseen requests (Phase~4). Each round starts from the previous round's final system, and a new deployment from stock plus memory entries, so the loop is recursive.

\subsection{Where Speculative Decoding Loses Speed}

\label{sec:model}

Phase~0 locates speed losses with a cost model that expresses the speedup over AR decoding as
\begin{equation}
S(N) \;=\; \frac{\tau(N)}{c(N)}, \qquad c(N) \;=\; \frac{T_{\mathrm{step}}(N)}{T_{\mathrm{AR}}},
\label{eq:model}
\end{equation}
where $\tau(N)$ is the number of tokens a step accepts verifying $N$ rows (one per node with one request), $T_{\mathrm{step}}(N)$ the step time with drafting and $T_{\mathrm{AR}}$ one AR step on the same engine. Unless otherwise stated, single-request numbers in this paper use DFlash2 and Qwen3.8-27B on an H100 GPU.\looseness=-1

\runin{Unused candidates ($\tau$).}Stock accepts far fewer tokens than its drafter's candidates allow. In \emph{exact replay}, which reruns tree selection and verification offline on recorded drafter lattices and target outputs, stock DFlash2's 8-position chain accepts $\tau=4.25$ tokens per step (4.65 live on the bench requests of \S\ref{sec:setup}). An oracle that keeps accepting the target's greedy token while it stays in the drafter's top-16 would accept 6.42 at 8 positions and 8.20 at 16 (\S\ref{sec:a-tau}).

\runin{Weight streaming ($c$, single request).}With one request, reading the target's weights dominates the step, so extra verify rows are nearly free. Before the first round, GEMM time grew only from 17.88\,ms for one row to 18.58\,ms for 64. After the first round, reading target weights (linears and LM head) took 80\% of the measured 23.61\,ms step. Even at the H100's peak 3.35\,TB/s, reading them would take 92\% of the time needed to read all bytes of the step (Figure~\ref{fig:motiv}a).

\runin{Per-row cost ($c$, batching).}Each batched request adds recurrent-state and attention cost, and GEMM time steps up at each tile boundary. Each new prompt's prefill streams the weights again. Stock DFlash2's speedup over AR in FP8 falls from 3.40$\times$ with one request to 1.89$\times$ with 24, and on Qwen3.5-4B from 3.73$\times$ to 1.49$\times$ with 32 (Figure~\ref{fig:motiv}b).

\runin{Ceilings.}To rank these effects, the cost model computes a \emph{ceiling} for the system a round starts from. The ceiling is the speedup over that system a step could reach if it drafted once, verified the best tree in one exact pass and ran every kernel class (target GEMMs, recurrent layers, attention, drafter, prefill and the rest) at its \emph{byte floor}, defined as the larger of its FLOP time at 60\% of peak dense throughput and its byte time at 90\% of peak bandwidth. The ceiling factors as
\begin{equation}
\frac{\tau_{\mathrm{best}}}{\tau}\cdot\frac{T_{\mathrm{prof}}}{T_{\mathrm{floor}}}\cdot\frac{T_{\mathrm{step}}}{T_{\mathrm{prof}}},
\label{eq:ceiling}
\end{equation}
where $\tau$ and $T_{\mathrm{step}}$ are measured on the starting system, and $\tau_{\mathrm{best}}$ is the $\tau$ of the best-first tree in exact replay at the best node budget. The $\tau$ factor measures how many more tokens a better tree would accept. The kernel factor compares the profiled step time $T_{\mathrm{prof}}$ with the sum of its byte floors $T_{\mathrm{floor}}$, measuring slow kernels. The overhead factor compares $T_{\mathrm{step}}$ with $T_{\mathrm{prof}}$, measuring time outside the profile.

\runin{Ceilings in use.}Phase~0 of each round recomputes the factors from recorded lattices and profiles, and the largest factor identifies the gap. \S\ref{sec:a-rounds} traces them over three rounds.

\subsection{Agents and Phases}

\label{sec:round}

\label{sec:agent}

SpecPilot separates the agents that modify the system from the controller that measures it (Figure~\ref{fig:agent}). A \emph{master agent} plans \emph{directions}, one idea per GPU, each with an estimated \emph{headroom} (expected gain). A \emph{sub-agent} implements each direction as \emph{patches}, code changes behind options whose defaults reproduce stock. A patch must pass correctness checks and token forcing (\S\ref{sec:forcing}) on \emph{development} requests, which agents see. The \emph{controller}, scripts that agents call but cannot modify, times it against the current best in \emph{paired runs}, where both systems serve the same requests on two GPUs, swapping GPUs between runs. It \emph{promotes} the patch only if its gain's 95\% lower bound exceeds 1\% plus twice the stock-versus-stock noise ($<$ 0.1\% for one request on H100), or ${\sim}$3\% if its outputs diverge from the current best's.

\runin{Phase 1: Exploration.}The master starts from Phase~0's cost model and proposes directions aimed at its largest factors. A single direction can stall on an idea that does not pay off, so the master explores the most promising few at once. A sub-agent first checks the headroom with the cheapest evidence, such as exact replay, a microbenchmark or a timing run on one GPU. It builds a new algorithm only if the headroom, net of runtime overhead, is at least 3\%. It closes a direction only after measuring the implemented mechanism, unless that would take over 10 GPU-minutes.\looseness=-1

\runin{Phases 2--3: Integration and kernel tuning.}Many patches gain too little to promote alone, but enough together. An integration sub-agent stacks them on the current best and times the stack under the same rule. A kernel sub-agent profiles it and tunes kernels furthest from byte floors.\looseness=-1

\runin{Phase 4: Sealed final.}Each round ends with a \emph{sealed final}, which times the system against stock on unseen requests and reruns token forcing on its outputs and on longer outputs of up to 4096 tokens. A final that fails either check is rejected.

\begin{figure*}[t]

\centering

\includegraphics[trim=6 7 6 3,clip,width=\textwidth]{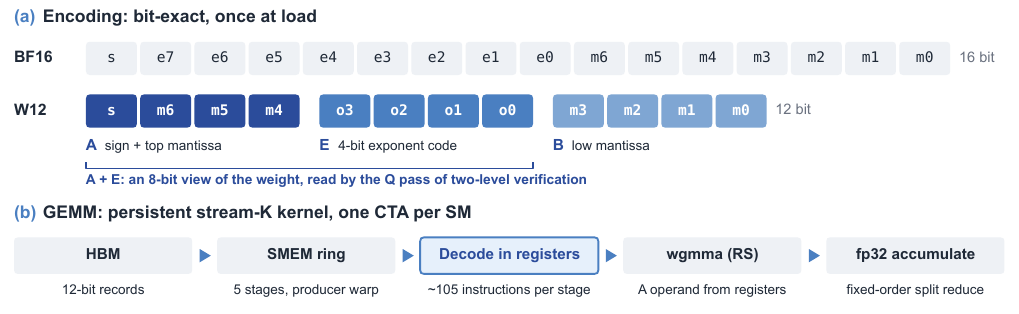}

\vskip -0.05in

\caption{W12. \textbf{(a)}~Each BF16 weight becomes three 4-bit planes: A holds the sign and the top three mantissa bits, E a 4-bit exponent offset within a 16-binade window per output channel, and B the low mantissa bits. A and E form the 8-bit view that the Q pass of two-level verification reads. The remaining weights, a fraction of $7.3\times10^{-5}$ on Qwen3.8-27B, go to an exact escape list. \textbf{(b)}~The GEMM decodes W12 records in registers and feeds them to \texttt{wgmma} as the register operand.}

\label{fig:w12tlv}

\vskip -0.1in

\end{figure*}

\runin{A worked example.}In round two, step cost led (\S\ref{sec:a-rounds}), so the master opened a direction cutting weight bytes (W12, Figure~\ref{fig:w12tlv}). Byte arithmetic put its headroom at 25\% of the 15.30\,ms weight read, and a microbenchmark closed a denser 11-bit variant that decoded 3$\times$ slower. The first patch was promoted at 1.076$\times$ in paired runs and a revised kernel gave 1.063$\times$. The DSpark, RTX PRO 6000 and Muse-Glimmer deployments reused W12 from memory (\S\ref{sec:general}).

\begin{figure}[t]
\centering
\includegraphics[width=0.6\textwidth]{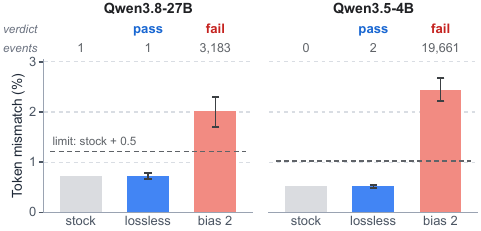}
\vskip -0.08in
\caption{Token forcing on lossless and faulty systems in sealed finals. Bars give the mismatch rate with 95\% CIs, and dashed lines the rate-rule limit. Above the bars are the verdict and the count of large-margin events. Faulty systems add a logit bias of 2 to every 7th vocabulary id after 2048 tokens.}
\label{fig:forcing}
\vskip -0.12in
\end{figure}

\subsection{Token Forcing}

\label{sec:forcing}

Losslessness holds by construction (\S\ref{sec:built}). Token forcing tests the implementation against stock for bugs and reward hacking. A teacher-forced \emph{scorer} feeds the system's outputs back through the target on the stock engine with an FP32 LM head. A token is a \emph{mismatch} if it differs from the scorer's argmax, and its \emph{margin} is the log-probability gap to that argmax. Stock mismatches on 0.72\% of tokens, almost all near-ties from kernel rounding. Phase~0 sets the near-tie threshold $\delta$ by rounding up the largest margin at stock's first departures from AR to a multiple of 1/8.

A system is rejected under either of two rules. The \emph{rate rule} rejects it if its mismatch rate exceeds stock's by more than 0.5 points at 95\% confidence. The \emph{event rule} rejects it if a binomial test finds its unexplained \emph{large-margin events}, mismatches with margin above $2\delta$, more frequent than stock's. An audit explains an event if stock reproduces it on its own paths or margins, or if the scorer's argmax changes with the prefill length. Figure~\ref{fig:forcing} tests lossless and faulty systems.


\subsection{Structured Memory}

\label{sec:memory}

Memory lets a new deployment's first round start from stock plus validated entries chosen by the master. After a sealed final, each promoted mechanism becomes one memory \emph{entry}, holding a code delta against the entries it requires, a registry record of its variants, correctness contract and options, and a summary of its evidence, limits and untested regimes. An entry is kept only if applying it with its requirements to stock reproduces its source system's code exactly. A reused entry is re-tuned and re-timed in the new deployment and never counts as a new discovery. The memory now holds 140 entries with up to twelve requirements each and covers every setting in \S\ref{sec:setup} (15 entries specific to sm\_120).

\section{Discovered Lossless Algorithms}

\label{sec:built}

We describe each mechanism and why it is lossless.

\subsection{W12: A Lossless 12-bit Weight Format}

\label{sec:w12}

W12 reduces the step cost relative to AR decoding by reading fewer bytes per weight while keeping every bit exact. Most BF16 weights in an output channel span only a few binades (powers of two). For each channel $n$, W12 selects the 16-binade window $[E_n-15, E_n]$ that covers the most weights. Each weight with in-window exponent $e$ is stored as its sign bit, a 4-bit exponent offset $e-(E_n-15)$, and its seven mantissa bits, for a total of 12 bits (Figure~\ref{fig:w12tlv}a). Weights outside the window, zeros and denormals become \emph{escapes}, which a sparse per-channel list restores exactly. W12 averages only 12.01--12.02 bits per weight on eight dense models and shrinks Qwen3.8-27B's converted layers from 44.67 to 33.58\,GB. At load time, the server checks every decoded tensor against the checkpoint bit for bit.

The GEMM kernel decodes W12 in registers, so the memory stream hides decoding. Each 12\,KB record covers a $128{\times}64$ weight tile. It holds the \emph{8-bit view}, one byte per weight with the sign, top three mantissa bits and exponent offset (planes A and E), then the low mantissa bits (plane B). A stream-K kernel (Figure~\ref{fig:w12tlv}b) splits the reduction loop over thread blocks in a fixed order, so the GEMM is deterministic.\looseness=-1

\runin{Certified LM head.} On H100, W12 leaves the LM head in BF16. An INT8 copy with a per-row Cauchy--Schwarz error bound discards every token that provably cannot be the argmax, and the survivors are rescored exactly. When too many survive, the full head runs instead. The argmax thus never changes. This reduces LM head latency from 878 to 482\,$\mu$s, outperforming W12. On the RTX PRO 6000, the certified head is slower than BF16 (2598 vs.\ 1750\,$\mu$s), so W12 converts the LM head there.\looseness=-1


\subsection{Tree Verification for Hybrid Targets}

\label{sec:trees}

\begin{figure}[t]
\centering
\includegraphics[trim=3.5 0.5 254 19.5,clip,scale=1.12]{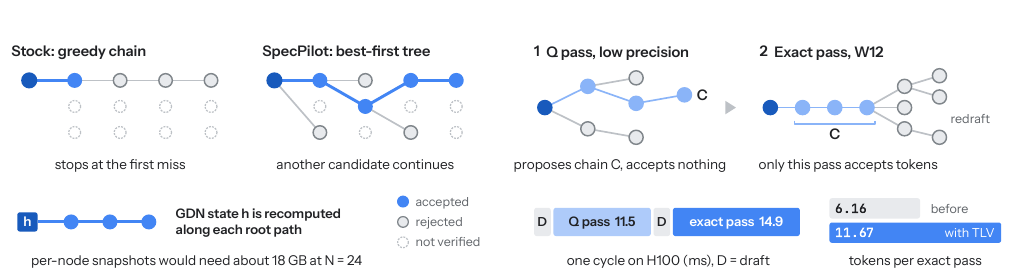}\panelrule{116pt}
\includegraphics{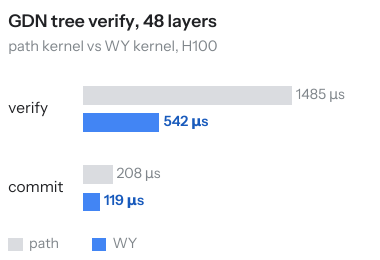}
\vskip -0.05in
\caption{Best-first tree on a hybrid target. On the same lattice, stock verifies only the greedy chain, which stops at its first miss. The best-first tree keeps the $N$ nodes with the highest path score $\sum \log p(\mathrm{child} \mid \mathrm{parent})$, so another candidate can continue. On hybrid targets, each root path recomputes its GDN state. The right panel compares the path and WY verify kernels over 48 GDN layers on H100.}
\label{fig:tree}
\vskip -0.04in
\end{figure}

Round one built best-first trees (Figure~\ref{fig:tree}) on the DFlash2 lattice, scoring each edge by $\log\mathrm{softmax}$ of selector scores given its parent, not by DDTree's marginal scores~\cite{ringel2026ddtree}. The drafter runs 16 positions, twice its trained block size, and the tree verifies $N{=}32$ nodes up to depth 15, more than the drafted positions (a \emph{wide tree}).

\runin{Recurrent state without snapshots.}SpecPilot recomputes the recurrent state along each root path instead of storing it per node. A path verify kernel walks each path from the committed state, and a commit kernel replays the accepted path into all 48 GDN layers, matching the stock kernel bit for bit on chains. The faster \emph{WY verify kernel}, following TreeWY~\cite{treewy2026}, turns the chain of rank-one updates in Eq.~(\ref{eq:gdn}) into matrix products, solving all paths at once on tensor cores. It splits the FP32 state into three BF16 parts that represent all 24 significand bits exactly, accumulates tensor-core products in FP32, and stays within one BF16 ulp of the path kernel's outputs. Both kernels change only rounding, so the exact pass stays exact.

\runin{DSpark lattice.}DSpark's Markov head scores adjacent-token pairs like the DFlash2 selector, so the best-first builder runs unchanged on a \emph{DSpark lattice} that scores each child by Eq.~(\ref{eq:markov}) with its parent taken as $x_{s-1}$.

\subsection{Two-Level Verification}

\label{sec:tlv}

\begin{figure}[t]
\centering
\includegraphics[trim=245 0.5 3.5 19.5,clip,scale=1.12]{figures/fig_tree_tlv_split.pdf}\panelrule{114pt}
\includegraphics{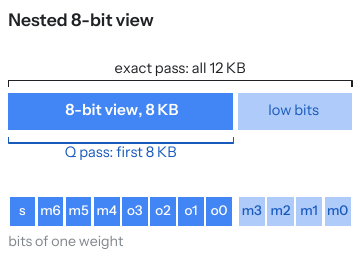}
\vskip -0.05in
\caption{Two-level verification. A low-precision Q pass checks the draft tree and returns chain C, the path that matches the Q pass's argmax plus a bonus token, without updating the target's state. The drafter drafts again from the end of C, and one exact pass over the W12 weights verifies C and the new subtree. Times are from one final H100 cycle, and tokens per exact pass from TLV's round. The right panel shows the 8-bit view nested in each W12 record, which the Q pass reads.}
\label{fig:tlv}
\vskip -0.04in
\end{figure}

Each TLV cycle has two passes (Figure~\ref{fig:tlv}). A \emph{Q pass} over the 8-bit view of W12 checks the draft tree and returns the path that matches its argmax and a bonus token as a chain $C$, without updating the KV cache or the recurrent state. The drafter drafts again from the end of $C$, and its best-first subtree, truncated to $N-|C|$ nodes, is attached below $C$. One exact pass over the W12 weights verifies the combined 32-node tree and accepts the longest matching path. TLV is lossless because only the exact pass accepts tokens. Under TLV, $\tau$ denotes tokens per exact pass and $T_{\mathrm{step}}$ is the time of the two-pass \emph{cycle}, including both drafting steps.

\runin{Zero-copy Q pass.}TLV does not maintain a separate low-precision target model, since it reads the 8-bit view already nested in W12. This view is the first two thirds of each W12 record. With the dropped mantissa bits set to their midpoint, the Q pass matches the exact pass's top-1 token at 98.7\% of positions, whereas the exact pass reads the full records.

\subsection{Batched Serving}

\label{sec:load}

Every batched row costs time (\S\ref{sec:model}). These mechanisms only pick the rows to draft and verify, so they are lossless.\looseness=-1

\runin{Shared node budget.}Each step verifies a fixed node budget shared across requests, taking the top nodes by path log-probability. As best-first trees are prefix-closed (\S\ref{sec:bg-trees}), each request's share is a valid tree. On the \emph{FP8 target} (Qwen3.8-27B in FP8) at 24 requests, the 192-node budget is on a \emph{plateau} of the target GEMM's cost, flat until the next tile.

\runin{Per-request drafter block.}Each request drafts 16, 12 or 8 positions, chosen on the GPU from its recent acceptance and draft confidence, and its tree depth is capped to match. This keeps the FP8 target's drafter within 256 rows, the end of the drafter GEMM's plateau, with 17 to 24 requests.\looseness=-1

\runin{Prompt admission.}A new prompt's rows enter the verify step, so one weight stream serves both. On the FP8 target, a prompt of up to 384 tokens enters as a chain of prompt rows, but only when the scheduler would otherwise hold it back. Both FP8 final systems admit prompts this way. For batched BF16 Qwen3.8-27B, prompts enter in 128-row segments that share the verify step's projection and MLP GEMMs.

\section{Evaluation}

\label{sec:eval}

We examine whether the speedups hold across drafters, targets, serving regimes and GPU platforms (\S\ref{sec:main}) and whether memory carries them to new deployments (\S\ref{sec:general}).

\subsection{Setup}

\label{sec:setup}

\runin{Hardware and agents.}Each SGLang~\cite{zheng2024sglang} server uses one H100 SXM5 or RTX PRO 6000 Blackwell Server Edition (sm\_120, 1.51\,TB/s measured), on a frozen v0.5.20 build, token-identical to the release. Agents are Claude Opus 5.5 in Claude Code~\cite{anthropic2025claudecode}.

\runin{Models and requests.}With one request, all targets run in BF16: Qwen3.8-27B~\cite{qwen2026qwen38} with DFlash2 on both GPUs and with DSpark on H100, and Muse-Glimmer-30B~\cite{meta2026museglimmer} with DFlash2 on the RTX PRO 6000. With batching on H100, Qwen3.8-27B runs with DFlash2 or DSpark in FP8, and with DFlash2 in BF16 at 6, 8 and 10 requests as GPU memory allows. Qwen3.5-4B~\cite{qwen2026qwen35} uses DFlash2 and Gemma-4-12B~\cite{gemmateam2026gemma4technicalreport} uses EAGLE-3, both in BF16. Requests are from GSM8K~\cite{cobbe2021gsm8k}, HumanEval~\cite{chen2021humaneval}, MATH500~\cite{hendrycks2021math,lightman2024verify}, MBPP~\cite{austin2021mbpp} and MT-Bench~\cite{zheng2023mtbench}. Decoding is greedy with thinking enabled, and batched clients run closed-loop with a fixed number of requests in flight (the \emph{batch size}).\looseness=-1


\runin{Metrics.}A gain is the geometric mean of per-request output-rate ratios per benchmark and batch size, combined by an equal-weight geometric mean. Every gain over stock comes from a sealed final (\S\ref{sec:agent}) with two paired runs. Every final system passed token forcing. Speedups over AR (Tables~\ref{tab:h100} and~\ref{tab:load}) come from interleaved runs of AR, stock and SpecPilot on development requests, within $-$2\% to +6\% of the sealed gains on unseen requests.

\begin{table*}[t]
\caption{Speedup over AR and $\tau$, the tokens accepted per exact target pass, with one request on BF16 targets. SpecPilot uses two-level verification, so each exact pass verifies two chained drafts and accepts far more tokens than stock.}
\label{tab:h100}\label{tab:pro6000}
\vskip 0.04in
\centering
\small
\setlength{\tabcolsep}{2.2pt}
\renewcommand{\arraystretch}{1.15}
\begin{adjustbox}{max width=\textwidth}
\begin{tabular}{ll|cccccccccc|cc}
\toprule
\multirow{2}{*}{Model} & \multirow{2}{*}{Method} & \multicolumn{4}{c}{\textsc{Math}} & \multicolumn{4}{c}{\textsc{Code}} & \multicolumn{2}{c|}{\textsc{Chat}} & & \\
\cmidrule(l){3-14}
 & & \multicolumn{2}{c}{GSM8K} & \multicolumn{2}{c}{MATH500} & \multicolumn{2}{c}{HumanEval} & \multicolumn{2}{c}{MBPP} & \multicolumn{2}{c|}{MT-Bench} & \multicolumn{2}{c}{\emph{Avg.}} \\
\midrule
\multicolumn{2}{l|}{\textit{H100 SXM5}} & \footnotesize Speedup & \footnotesize $\tau$ & \footnotesize Speedup & \footnotesize $\tau$ & \footnotesize Speedup & \footnotesize $\tau$ & \footnotesize Speedup & \footnotesize $\tau$ & \footnotesize Speedup & \footnotesize $\tau$ & \footnotesize Speedup & \footnotesize $\tau$ \\
\cmidrule(lr){1-14}
 & DFlash2 & 4.47$\times$ & 5.82 & 4.31$\times$ & 5.26 & 3.86$\times$ & 4.69 & 3.83$\times$ & 4.71 & 3.36$\times$ & 3.84 & 3.95$\times$ & 4.65 \\
\rowcolor{sprow}\cellcolor{white} & +SpecPilot & 10.33$\times$ & 16.69 & 10.00$\times$ & 15.46 & 8.49$\times$ & 13.05 & 8.24$\times$ & 12.66 & 7.55$\times$ & 10.81 & 8.86$\times$ & 13.08 \\
\cmidrule(l){2-14}
 & DSpark & 4.43$\times$ & 5.87 & 4.15$\times$ & 5.11 & 3.57$\times$ & 4.39 & 3.50$\times$ & 4.45 & 3.17$\times$ & 3.61 & 3.74$\times$ & 4.42 \\
\rowcolor{sprow}\cellcolor{white}\multirow{-4}{*}{Qwen3.8-27B} & +SpecPilot & 11.01$\times$ & 16.67 & 10.69$\times$ & 15.66 & 8.86$\times$ & 12.72 & 9.02$\times$ & 12.79 & 7.85$\times$ & 10.17 & 9.41$\times$ & 12.85 \\
\midrule
\multicolumn{2}{l|}{\textit{RTX PRO 6000}} & \footnotesize Speedup & \footnotesize $\tau$ & \footnotesize Speedup & \footnotesize $\tau$ & \footnotesize Speedup & \footnotesize $\tau$ & \footnotesize Speedup & \footnotesize $\tau$ & \footnotesize Speedup & \footnotesize $\tau$ & \footnotesize Speedup & \footnotesize $\tau$ \\
\cmidrule(lr){1-14}
 & DFlash2 & 5.23$\times$ & 6.82 & 4.81$\times$ & 6.07 & 4.47$\times$ & 5.62 & 4.24$\times$ & 5.25 & 3.72$\times$ & 4.06 & 4.46$\times$ & 5.23 \\
\rowcolor{sprow}\cellcolor{white}\multirow{-2}{*}{Qwen3.8-27B} & +SpecPilot & 12.83$\times$ & 16.75 & 11.99$\times$ & 15.02 & 10.47$\times$ & 13.24 & 10.55$\times$ & 12.98 & 8.93$\times$ & 10.25 & 10.87$\times$ & 12.80 \\
\cmidrule(lr){1-14}
 & DFlash2 & 5.08$\times$ & 5.31 & 5.64$\times$ & 6.70 & 4.91$\times$ & 5.77 & 4.59$\times$ & 5.51 & 3.50$\times$ & 3.76 & 4.69$\times$ & 5.16 \\
\rowcolor{sprow}\cellcolor{white}\multirow{-2}{*}{Muse-Glimmer-30B} & +SpecPilot & 11.47$\times$ & 13.24 & 12.85$\times$ & 16.02 & 10.75$\times$ & 13.69 & 10.25$\times$ & 13.08 & 8.34$\times$ & 9.80 & 10.63$\times$ & 12.64 \\
\bottomrule
\end{tabular}
\end{adjustbox}
\vskip -0.12in
\end{table*}

\begin{table*}[t]
\caption{Speedup over AR and $\tau$ with batching on H100, averaged over 16/20/24 requests on Qwen3.8-27B in FP8, and over 24/28/32 on Qwen3.5-4B and 16/32/48 on Gemma-4-12B, both in BF16.}
\label{tab:load}
\vskip 0.04in
\centering
\small
\setlength{\tabcolsep}{3.6pt}
\renewcommand{\arraystretch}{1.15}
\begin{adjustbox}{max width=\textwidth}
\begin{tabular}{ll|cccccccccc|cc}
\toprule
\multirow{2}{*}{Model} & \multirow{2}{*}{Method} & \multicolumn{4}{c}{\textsc{Math}} & \multicolumn{4}{c}{\textsc{Code}} & \multicolumn{2}{c|}{\textsc{Chat}} & & \\
\cmidrule(l){3-14}
 & & \multicolumn{2}{c}{GSM8K} & \multicolumn{2}{c}{MATH500} & \multicolumn{2}{c}{HumanEval} & \multicolumn{2}{c}{MBPP} & \multicolumn{2}{c|}{MT-Bench} & \multicolumn{2}{c}{\emph{Avg.}} \\
\midrule
\multicolumn{2}{l|}{\textit{H100 SXM5}} & \footnotesize Speedup & \footnotesize $\tau$ & \footnotesize Speedup & \footnotesize $\tau$ & \footnotesize Speedup & \footnotesize $\tau$ & \footnotesize Speedup & \footnotesize $\tau$ & \footnotesize Speedup & \footnotesize $\tau$ & \footnotesize Speedup & \footnotesize $\tau$ \\
\cmidrule(lr){1-14}
 & DFlash2 & 2.11$\times$ & 5.02 & 2.19$\times$ & 5.03 & 2.05$\times$ & 4.56 & 2.12$\times$ & 5.03 & 1.69$\times$ & 3.80 & 2.02$\times$ & 4.56 \\
\rowcolor{sprow}\cellcolor{white} & +SpecPilot & 4.39$\times$ & 6.14 & 4.33$\times$ & 5.95 & 3.81$\times$ & 5.38 & 4.07$\times$ & 6.20 & 3.07$\times$ & 4.33 & 3.90$\times$ & 5.39 \\
\cmidrule(l){2-14}
 & DSpark & 2.12$\times$ & 5.00 & 2.16$\times$ & 4.87 & 1.95$\times$ & 4.38 & 1.96$\times$ & 4.81 & 1.56$\times$ & 3.58 & 1.94$\times$ & 4.37 \\
\rowcolor{sprow}\cellcolor{white}\multirow{-4}{*}{Qwen3.8-27B} & +SpecPilot & 4.60$\times$ & 6.28 & 4.39$\times$ & 6.11 & 3.82$\times$ & 5.45 & 3.97$\times$ & 6.07 & 3.05$\times$ & 4.32 & 3.93$\times$ & 5.42 \\
\cmidrule(lr){1-14}
 & DFlash2 & 1.62$\times$ & 6.82 & 1.88$\times$ & 8.17 & 1.60$\times$ & 7.15 & 1.27$\times$ & 5.88 & 1.25$\times$ & 5.16 & 1.51$\times$ & 6.58 \\
\rowcolor{sprow}\cellcolor{white}\multirow{-2}{*}{Qwen3.5-4B} & +SpecPilot & 4.04$\times$ & 7.30 & 4.61$\times$ & 8.60 & 4.00$\times$ & 7.42 & 3.30$\times$ & 6.31 & 3.11$\times$ & 5.49 & 3.77$\times$ & 6.97 \\
\cmidrule(lr){1-14}
 & EAGLE-3 & 0.69$\times$ & 2.00 & 0.79$\times$ & 2.36 & 0.66$\times$ & 1.97 & 0.69$\times$ & 2.08 & 0.61$\times$ & 1.78 & 0.68$\times$ & 2.04 \\
\rowcolor{sprow}\cellcolor{white}\multirow{-2}{*}{Gemma-4-12B} & +SpecPilot & 2.28$\times$ & 2.82 & 2.41$\times$ & 2.94 & 2.22$\times$ & 2.64 & 1.88$\times$ & 2.28 & 1.74$\times$ & 2.02 & 2.09$\times$ & 2.54 \\
\bottomrule
\end{tabular}
\end{adjustbox}
\vskip -0.12in
\end{table*}

\runin{Baselines.}To keep the comparison fair, stock is tuned for each GPU and regime, using the pretrained drafter and the fastest upstream SGLang options we found. On H100, stock DFlash2 drafts its trained block size, as in the released configuration. On the RTX PRO 6000, the best stock for Qwen3.8-27B uses block 12 with Triton attention and is 9.9\% faster than the H100 configuration. No EAGLE-3 configuration beats the default by more than 2.4\%. For single requests on H100, stock DFlash2 with DDTree trees is a stronger baseline, at 4.82$\times$ over AR on the bench.\looseness=-1

\subsection{Main Results}

\label{sec:main}

With one request, SpecPilot is more than twice as fast as stock in every setting (Table~\ref{tab:sealed}), most on Qwen3.8-27B on the RTX PRO 6000. Over AR, it reaches 8.86--9.41$\times$ on H100 and 10.63--10.87$\times$ on the RTX PRO 6000 (Table~\ref{tab:h100}). On H100, it is also 1.84$\times$ faster than the DDTree baseline.

With batching on FP8 Qwen3.8-27B, SpecPilot beats stock at every batch size, by 1.85$\times$ with DFlash2 and 1.94$\times$ with DSpark overall, raising the speedup over AR from about 2$\times$ to 3.90$\times$ and 3.93$\times$ (Table~\ref{tab:load}). In BF16, it is 1.71$\times$ faster, at 5.67$\times$ over AR against 3.20$\times$ for stock. Targets with weaker stock gain more, 2.39$\times$ on Qwen3.5-4B and 3.11$\times$ on Gemma-4-12B, where stock trails AR.\looseness=-1

\subsection{Generality Through Memory}

\label{sec:general}

Memory generalizes across drafters, target models and GPU platforms, even to EAGLE-3, a sequential drafter. In each new deployment, the agents reused validated mechanisms and adapted them to the new drafter, model or GPU.

\runin{New drafters.}DSpark reused the tree, W12 and TLV entries, which one promotion applied together with the new DSpark lattice (2.110$\times$). DSpark also added the INT6 copy for the Q pass. EAGLE-3 on Gemma-4-12B reimplemented two designs from memory, W12 on the MLP and prompt admission. Its own entries attach an n-gram chain to the EAGLE tree, select nodes by acceptance probabilities calibrated on recorded trees and add single-KV-head attention kernels. Its first round reached a sealed 2.67$\times$, and promotions like calibrated selection and prompt admission raised the final speedup to 3.11$\times$.

\runin{A new GPU platform.}The RTX PRO 6000 has 99\,KB of shared memory per thread block and no \texttt{wgmma}. Porting the H100 tree and W12 entries, with W12 on \texttt{mma.sync} and tree verification on the Triton attention backend, alone gave a sealed 1.92$\times$ over stock, most of the final gain. TLV with split-KV verify attention, which splits each request's KV sequence across thread blocks, and an INT6/INT8 Q-pass copy raised it to 2.44$\times$. This exceeds its 2.12$\times$ ceiling over stock, which assumes one draft per pass (\S\ref{sec:model}).

\runin{A new model family.}Muse-Glimmer-30B, a dense model without recurrent layers, reused memory except the GDN entries. Its first promotion applied the tree, W12, TLV and the INT6/INT8 copy, with exact sliding-window attention in the 32-node tree. Kernel fusion with L2 prefetching then raised the final speedup to 2.21$\times$ on the RTX PRO 6000.

\section{Analysis}

\label{sec:analysis}

\subsection{How the System Evolved over Three Rounds}

\label{sec:a-rounds}

\begin{figure*}[t]
\centering
\includegraphics[width=0.88\textwidth]{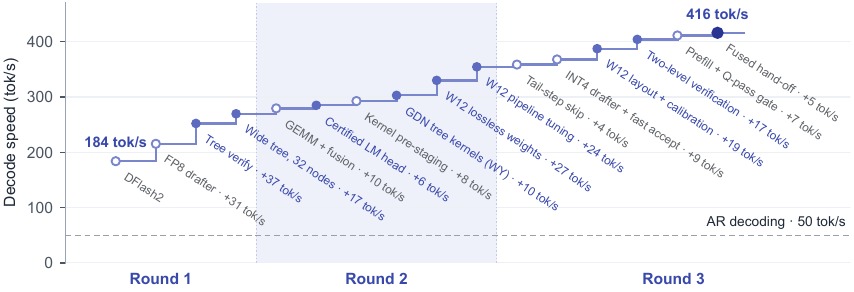}
\vskip -0.06in
\caption{Single-request Qwen3.8-27B with DFlash2 on H100 over three rounds. Filled points mark steps of the algorithms in \S\ref{sec:w12}--\ref{sec:tlv}.}
\label{fig:evolution}
\vskip -0.1in
\end{figure*}

The system went from stock's 184 to 416 tokens/s in three rounds (Figure~\ref{fig:evolution}), and each round faced a different bottleneck. The cost model re-measured the ceiling each round, so the master redirected its parallel search to the factor holding the most speed instead of refining the previous gain.\looseness=-1

\runin{Round one: tokens.}Stock verified one chain though extra rows were nearly free, and the $\tau$ factor led the ceiling. The round built best-first trees (\S\ref{sec:trees}) and reached 270 tokens/s. Exact replay showed a better tree on this lattice would add little, which moved the bottleneck from tokens to the step.

\runin{Round two: weight bytes.}Reading target weights took 80\% of the step, so the bytes per weight became the largest lever. The master opened a direction on a lossless 12-bit format (\S\ref{sec:w12}) and kept it through three promotions across two rounds. Its value came from the kernel as much as the format, since the W12 GEMM decodes weights in registers and beats cuBLAS at the few rows of a verify pass (\S\ref{sec:a-w12}).

\runin{Round three: the unit of a step.}The W12 \emph{single-pass step}, with one exact pass, ran about 25\% above its byte floors, and trees were near their best on the lattice, so neither factor had much headroom. The remaining gain lay in an assumption of the ceiling, one exact pass per draft. TLV removed it by drafting twice per exact pass (\S\ref{sec:tlv}), which nearly doubled tokens per exact pass and took the system to 416 tokens/s.

\runin{Single-agent baseline.}We compare SpecPilot with a single-agent loop that lacks the cost model, in the same setting. After three rounds, this baseline reached only 1.34$\times$ over stock, compared with 2.13$\times$ for SpecPilot. Without ceilings to rank the gaps, it focused on the most visible one, tokens per step, adding a 24-node best-first tree, a longer drafter block and cheaper drafter weights. It never addressed the target weight reads, although they took 80\% of the step. In contrast, SpecPilot recomputed the ceiling every round and followed the largest factor as it shifted, from trees to weight bytes (W12) and then to the unit of a step (TLV).

\subsection{Testing Token Forcing with Injected Faults}

\label{sec:a-forcing}

Token forcing separates faulty from lossless systems, but task accuracy does not. Lossless Qwen3.8-27B and 4B systems and two faulty ones, which add a logit bias of 2 to every 7th vocabulary id after 2048 tokens, ran sealed finals. Development outputs stop at 2048 tokens, so only the sealed finals' outputs of up to 4096 tokens reach the fault. On Qwen3.8-27B, accuracy changed by $-$0.3 points [$-$1.6, +0.7] for the lossless system and $-$1.6 [$-$3.6, +0.3] for the faulty one, so the intervals overlap. Token forcing accepts both lossless systems, whose mismatch rates (0.73\% and 0.52\%) match stock's (0.72\% and 0.53\%), and rejects both faulty ones, at 2.0\% and 2.4\%, with 192 and 191 of their 200 audited large-margin events left unexplained.

\subsection{W12 as a General Lossless Weight Format}

\label{sec:a-w12}

W12 is a lossless weight format useful beyond speculative decoding. Every dense model we encoded, from six families, compresses to about 12 bits per weight (\S\ref{sec:w12}), a quarter fewer bytes in every converted layer, without changing any output. Its GEMM speeds up any matrix product bound by weight bytes, which includes the single row of AR decoding as well as the few rows of a verify pass.\looseness=-1

\begin{figure*}[t]
\centering
\includegraphics[width=\textwidth]{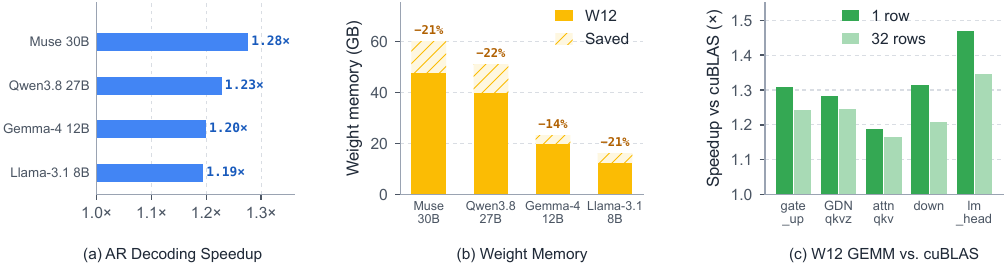}
\vskip -0.06in
\caption{W12 beyond speculative decoding, on H100. (a) AR decoding speedup over BF16 weights in SGLang with one request. (b) Weight memory in BF16 and W12. (c) W12 GEMM speedup over cuBLAS on Qwen3.8-27B projections at 1 and 32 rows, with cold L2.}
\label{fig:w12}
\vskip -0.1in
\end{figure*}

In AR decoding on SGLang, without speculation, W12 alone gives a 1.19--1.28$\times$ speedup on Muse-Glimmer-30B, Qwen3.8-27B, Gemma-4-12B and Llama-3.1-8B with one request (Figure~\ref{fig:w12}a), and 1.13--1.20$\times$ with 32 requests. It shrinks the weights by 14--22\% (Figure~\ref{fig:w12}b), which grows the KV pool 2.0$\times$ on Qwen3.8-27B and 2.5$\times$ on Muse-Glimmer-30B, leaving room for more requests.\looseness=-1

\newcommand{\sghead}[1]{\multicolumn{4}{l}{\textit{#1}} \\ \cmidrule(lr){1-4}}
\begin{table}[t]
\caption{Sealed gains over stock on unseen requests, and tokens per step from Tables~\ref{tab:h100} and~\ref{tab:load}.}
\label{tab:sealed}
\vskip 0.04in
\centering
\scriptsize
\setlength{\tabcolsep}{5pt}
\renewcommand{\arraystretch}{1.05}
\begin{tabular}{llrc}
\toprule
Target & Drafter & Gain & Tokens/step \\
\midrule
\sghead{Single request, H100 SXM5}
Qwen3.8-27B & DFlash2 & 2.13$\times$ & 4.65 $\to$ 13.08 \\
Qwen3.8-27B & DSpark & 2.38$\times$ & 4.42 $\to$ 12.85 \\
\midrule
\sghead{Single request, RTX PRO 6000}
Qwen3.8-27B & DFlash2 & 2.44$\times$ & 5.23 $\to$ 12.80 \\
Muse-Glimmer-30B & DFlash2 & 2.21$\times$ & 5.16 $\to$ 12.64 \\
\midrule
\sghead{Batching, H100 SXM5}
Qwen3.8-27B (FP8) & DFlash2 & 1.85$\times$ & 4.56 $\to$ 5.39 \\
Qwen3.8-27B (FP8) & DSpark & 1.94$\times$ & 4.37 $\to$ 5.42 \\
Qwen3.5-4B & DFlash2 & 2.39$\times$ & 6.58 $\to$ 6.97 \\
Gemma-4-12B & EAGLE-3 & 3.11$\times$ & 2.04 $\to$ 2.54 \\
\bottomrule
\end{tabular}
\vskip -0.12in
\end{table}

The shipped W12 GEMM beats cuBLAS on every Qwen3.8-27B projection, by 1.19--1.47$\times$ at one row, as in AR decoding, and by 1.17--1.35$\times$ at 32 rows, as in a verify pass (Figure~\ref{fig:w12}c). The first W12 kernel took 0.80--0.89 of cuBLAS's time at 32 rows. On the RTX PRO 6000, W12 is 1.18--1.68$\times$ faster on all 13 shapes at up to 32 rows. A 128-byte swizzled activation layout allows 48-row tiles, so W12 GEMMs stay near their byte floor and flat across verify rows.


\subsection{Tree Gains in Exact Replay}

\label{sec:a-tau}

\begin{figure}[t]
\centering
\includegraphics[width=0.56\textwidth]{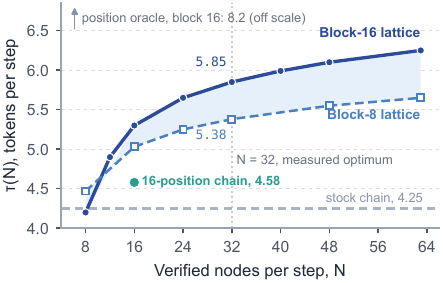}
\vskip -0.08in
\caption{Tokens per step $\tau$ in exact replay of single-request DFlash2 on Qwen3.8-27B, against the number of verified nodes $N$. Best-first trees run on lattices drafted with 8 and 16 positions. The gray line is stock's 8-position chain, and the oracle on the 16-position lattice reaches 8.2 tokens per step.}
\label{fig:tau}
\vskip -0.12in
\end{figure}

Exact replay reruns tree selection and verification on recorded lattices, so it isolates the tree from timing noise (Figure~\ref{fig:tau}). On the 8-position lattice, best-first trees raise $\tau$ from 4.25 (stock chain) to 5.38 at 32 nodes. Drafting 16 positions lifts the curve to 5.85 at 32 nodes and 6.25 at 63. At 16 nodes, best-first trees match an EAGLE-2-style top-4 tree (5.03 vs.\ 5.02) and beat a Sequoia tree (4.71), so further gains must come from the lattice and node budget.

\subsection{Why Mechanisms Depend on the Regime}

\label{sec:a-regime}

Mechanisms that spend more verify work per exact pass help with one request, where the weight stream hides that work, and hurt with batching, where it adds up. Wide trees, W12 and TLV gained with one request on all three Qwen3.8-27B deployments. With batching, 24-node trees lost 14.9\% on the FP8 target, and on BF16 Qwen3.8-27B a 24-position block lost 9.9\% and TLV 13\%. TLV's Q pass cuts bytes only in the attention and recurrent-layer projections. With the Q pass at 0.80--0.88 of a step, a cycle must accept nearly twice the tokens of a single-pass step to break even.\looseness=-1

Mechanisms that choose rows selectively gain with batching. With DSpark, the shared node budget accepted 6--14\% more tokens than equal-size per-request trees, and per-request drafter blocks gave 1.040$\times$. Prompt rows are cheap only if compute idles in the weight stream, so admitting only waiting prompts gained 3.5\% (unpaired) at 24 requests, while admitting every short prompt lost.\looseness=-1

\section{Related Work}
\label{sec:related}

\runin{Agents for systems.}LLM program search improves code with automated evaluators, as in FunSearch~\cite{romeraparedes2024funsearch}, AlphaEvolve~\cite{novikov2025alphaevolve} and ADRS~\cite{cheng2025adrs}, after AlphaDev~\cite{mankowitz2023alphadev} and search-based compilers such as Ansor~\cite{zheng2020ansor}. KernelBench~\cite{ouyang2025kernelbench} scores generated GPU kernels, and CUDA-L1~\cite{li2025cudal1} optimizes them with reinforcement learning.\looseness=-1

\runin{Drafters and trees.}Medusa~\cite{cai2024medusa}, EAGLE~\cite{li2024eagle,li2024eagle2,li2025eagle3}, Lookahead~\cite{fu2024lookahead} and self-speculative methods~\cite{zhang2024draftverify,elhoushi2024layerskip} change drafting. Sequoia~\cite{chen2024sequoia}, EAGLE-2~\cite{li2024eagle2}, OPT-Tree~\cite{wang2024opttree}, DDTree~\cite{ringel2026ddtree} and CaDDTree~\cite{caddtree2026} shape trees for acceptance or throughput, and the concurrent PCTree~\cite{pctree2026} builds trees from DSpark's Markov head. Our best-first trees score each edge given its parent, unlike DDTree, and verify more nodes than drafted positions. With batching, requests share one node budget. STree~\cite{wu2025stree} and TreeWY~\cite{treewy2026} verify hybrid-target trees without per-node state, and we add a commit that replays the accepted path.\looseness=-1

\runin{Hierarchical speculation.}Staged speculative decoding~\cite{spector2023staged}, TriForce~\cite{sun2024triforce} and ML-SpecQD~\cite{georganas2025mlspecqd} place an intermediate model between the drafter and the target, and QSpec~\cite{zhao2024qspec} and QuantSpec~\cite{tiwari2025quantspec} draft with a low-precision version of the target. In these methods, the intermediate model serves as a drafter. TLV instead uses a low-precision target as an intermediate verifier. It checks the drafter's full tree without accepting tokens, and one exact pass verifies its matching path plus a second draft (\S\ref{sec:tlv}).\looseness=-1

\runin{Speculation with batching.}TurboSpec~\cite{liu2024goodput} adapts to goodput, MagicDec~\cite{chen2024magicdec} handles large batches, and D-cut~\cite{dcut2026} prunes drafts across requests, while our node budget at a GEMM plateau takes the most probable nodes of all requests.\looseness=-1

\runin{Lossless weight compression.}ZipNN~\cite{hershcovitch2024zipnn} and DFloat11~\cite{zhang2025dfloat11} entropy-code weights to save memory but decode them before the GEMM, and DFloat11 runs 4.8--6.7$\times$ slower than cuBLAS on H100. ZipServ~\cite{fan2026zipserv} and Unweight~\cite{nikulin2026unweight} fuse decoding into the GEMM but do not consistently beat cuBLAS on HBM GPUs. W12 gives every weight a fixed 12-bit record that the GEMM decodes in registers. To our knowledge, it is the first lossless format whose GEMM beats cuBLAS on H100, by 1.17--1.47$\times$ from 1 to 32 rows.\looseness=-1

\section{Conclusion}
\label{sec:conclusion}

This work introduces SpecPilot, an autonomous agentic loop for algorithm discovery and systems optimization in lossless speculative decoding. Through ceiling-guided discovery, SpecPilot adapts to the distinct bottlenecks of complex serving regimes. We evaluate SpecPilot across four target models, three drafters, two GPU platforms and five benchmarks. Without training or quality loss, it raises the speedup over AR from stock's 4.46$\times$ to 10.87$\times$ with one request, and from 1.94$\times$ to 3.93$\times$ with batching.\looseness=-1

\clearpage

\begingroup
\setlength{\bibsep}{1pt plus 0.5pt}
\bibliography{refs}
\endgroup

\setcounter{topnumber}{2}
\clearpage
\appendix
\setcounter{figure}{0}\setcounter{table}{0}
\renewcommand{\thefigure}{\textbf{A\arabic{figure}}}
\renewcommand{\thetable}{\textbf{A\arabic{table}}}
\begin{figure*}[t]
\centering
\includegraphics[width=\textwidth]{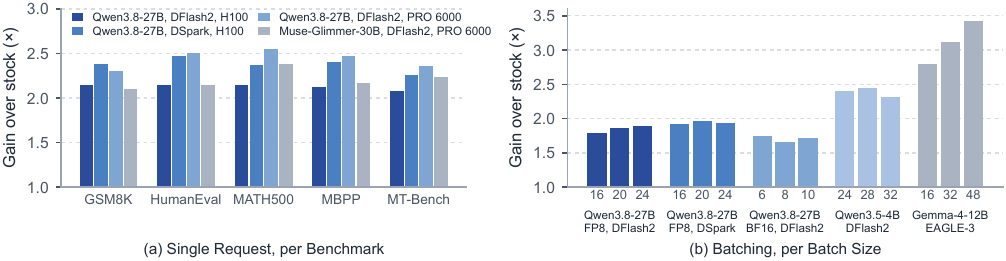}
\vskip -0.06in
\caption{Sealed gains over stock on unseen requests. (a) Single request, per benchmark. (b) Batching on H100, per batch size.}
\label{fig:app-results}
\vskip -0.1in
\end{figure*}

\section{Per-Benchmark and Per-Level Gains}

\label{app:results}

Every sealed final beats stock on every benchmark and at every batch size (Figure~\ref{fig:app-results}). With one request, the gain ranges from 2.08$\times$ on MT-Bench with DFlash2 on H100 to 2.55$\times$ on MATH500 with DFlash2 on the RTX PRO 6000. With batching, the gain on Gemma-4-12B grows with the batch size, from 2.80$\times$ at 16 requests to 3.43$\times$ at 48, where EAGLE-3's stock slows down most.

\begin{figure*}[t]
\centering
\includegraphics[width=\textwidth]{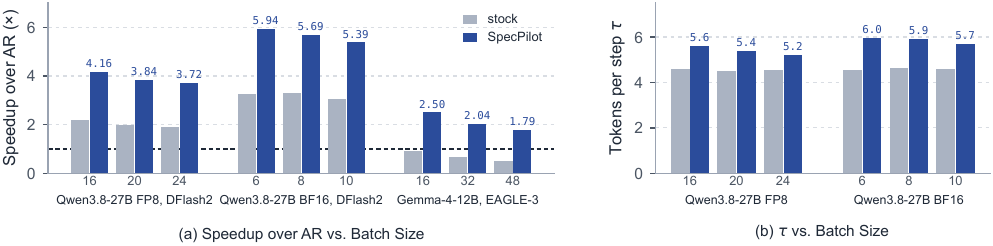}
\vskip -0.06in
\caption{Batching on H100, per batch size, measured on the interleaved bench. (a) Speedup over AR of stock and SpecPilot. (b) Tokens per step $\tau$ of stock and SpecPilot.}
\label{fig:app-load}
\vskip -0.1in
\end{figure*}

The speedup over AR stays high as the batch size grows (Figure~\ref{fig:app-load}a). On FP8 Qwen3.8-27B, SpecPilot is 4.16$\times$ faster than AR at 16 requests and 3.72$\times$ at 24, against 2.20$\times$ and 1.89$\times$ for stock. On BF16 Qwen3.8-27B it reaches 5.94$\times$ at 6 requests and 5.39$\times$ at 10. SpecPilot also accepts more tokens per step than stock with batching, 5.2--5.6 against 4.5--4.6 on the FP8 target and 5.7--6.0 against about 4.6 in BF16 (Figure~\ref{fig:app-load}b).

\begin{table}[t]
\centering
\begin{minipage}[t]{203.0pt}\vspace{0pt}
\caption{Decoding rates on the interleaved bench, in output tokens per second per request. Batched rows use Qwen3.8-27B in FP8 with DFlash2 on H100.}
\label{tab:app-rates}
\end{minipage}\hspace{36pt}%
\begin{minipage}[t]{182.2pt}\vspace{0pt}
\caption{W12 encoding of dense BF16 checkpoints. Every tensor decodes bit for bit to the original weights.}
\label{tab:app-w12-enc}
\end{minipage}

\vskip 0.03in
\begin{minipage}[t]{203.0pt}\vspace{0pt}
\scalebox{0.90}{%
\footnotesize%
\setlength{\tabcolsep}{3pt}%
\begin{tabular}{@{}lccc@{}}%
\toprule%
Setting & AR & Stock & SpecPilot \\%
\midrule%
\multicolumn{4}{@{}l}{\textit{Single request}} \\%
Qwen3.8-27B, DFlash2, H100 & 49.9 & 183.9 & 416.1 \\%
Qwen3.8-27B, DSpark, H100 & 49.9 & 172.7 & 435.3 \\%
Qwen3.8-27B, DFlash2, PRO 6000 & 26.9 & 108.2 & 272.3 \\%
Muse-Glimmer-30B, PRO 6000 & 25.7 & 110.6 & 255.6 \\%
\midrule%
\multicolumn{4}{@{}l}{\textit{Batching}} \\%
16 requests & 64.6 & 135 & 243 \\%
20 requests & 63.4 & 121 & 224 \\%
24 requests & 61.4 & 115 & 215 \\%
\bottomrule%
\end{tabular}}
\end{minipage}\hspace{36pt}%
\begin{minipage}[t]{182.2pt}\vspace{0pt}
\scalebox{0.90}{%
\renewcommand{\arraystretch}{1.146}%
\footnotesize%
\setlength{\tabcolsep}{3pt}%
\begin{tabular}{@{}lcc@{}}%
\toprule%
Model & Escape rate & Bits/weight \\%
\midrule%
Llama-3.1-8B-Instruct & $7.66\times10^{-5}$ & 12.015 \\%
Mistral-7B-Instruct-v0.3 & $1.21\times10^{-4}$ & 12.018 \\%
DeepSeek-LLM-7B-Chat & $7.60\times10^{-5}$ & 12.016 \\%
Gemma-4-12B & $8.71\times10^{-5}$ & 12.016 \\%
Gemma-4-31B & $7.89\times10^{-5}$ & 12.014 \\%
Muse-Glimmer-30B & $6.92\times10^{-5}$ & 12.013 \\%
Qwen3.5-4B & $7.88\times10^{-5}$ & 12.020 \\%
Qwen3.8-27B & $7.33\times10^{-5}$ & 12.014 \\%
\bottomrule%
\end{tabular}}
\end{minipage}
\end{table}

In absolute terms, SpecPilot decodes 416--435 tokens/s for one request on H100 and 256--272 on the RTX PRO 6000, against about 50 and 26 for AR (Table~\ref{tab:app-rates}). With 24 FP8 requests, each still decodes at 215 tokens/s, against 115 tokens/s for each stock request.

\section{Promotions on H100}

\label{app:chain}

\begin{table}[t]
\centering
\begin{minipage}[t]{165.0pt}\vspace{0pt}
\caption{Every promotion in the final lineage of single-request DFlash2 on Qwen3.8-27B (H100), in order, with its paired gain over the system it replaced. The sealed final is 2.13$\times$ over stock.}
\label{tab:app-chain}
\end{minipage}\hspace{36pt}%
\begin{minipage}[t]{228.7pt}\vspace{0pt}
\caption{Cost of each completed round: wall-clock hours, GPU-hours recorded by the controller, code changes the agents created, and promotions. 27B denotes Qwen3.8-27B. Rows use the DFlash2 drafter unless noted otherwise.}
\label{tab:app-cost}
\end{minipage}

\vskip 0.03in
\begin{minipage}[t]{165.0pt}\vspace{0pt}
\scalebox{0.90}{%
\footnotesize%
\setlength{\tabcolsep}{3pt}%
\begin{tabular}{@{}clc@{}}%
\toprule%
Round & Change & Gain \\%
\midrule%
1 & FP8 drafter, block-12 chain & 1.160$\times$ \\%
1 & Tree verification (16 nodes) & 1.160$\times$ \\%
1 & Wide tree (32 nodes) & 1.065$\times$ \\%
\midrule%
2 & Skinny GEMM and fusion & 1.031$\times$ \\%
2 & Certified LM head & 1.017$\times$ \\%
2 & Kernel pre-staging & 1.023$\times$ \\%
2 & GDN tree kernels (WY) & 1.030$\times$ \\%
2 & W12 & 1.076$\times$ \\%
2 & W12 pipeline tuning & 1.063$\times$ \\%
\midrule%
3 & Tail-step skip & 1.010$\times$ \\%
3 & INT4 drafter, fast accept & 1.022$\times$ \\%
3 & W12 layout, edge calibration & 1.045$\times$ \\%
3 & Two-level verification & 1.037$\times$ \\%
3 & Prefill graphs, Q-pass gate & 1.015$\times$ \\%
3 & Fused draft-to-exact hand-off & 1.010$\times$ \\%
\bottomrule%
\end{tabular}}
\end{minipage}\hspace{36pt}%
\begin{minipage}[t]{228.7pt}\vspace{0pt}
\scalebox{0.90}{%
\renewcommand{\arraystretch}{1.100}%
\footnotesize%
\setlength{\tabcolsep}{3pt}%
\begin{tabular}{@{}lcccc@{}}%
\toprule%
Setting & Hours & GPU-h & Changes & Promoted \\%
\midrule%
\multicolumn{5}{@{}l}{\textit{Single request}} \\%
27B, H100, round 1 & 11.7 & 28.3 & 26 & 4 \\%
27B, H100, rounds 2--3 & 19.3 & 57.3 & 103 & 13 \\%
27B DSpark, H100 & 13.6 & 27.3 & 25 & 4 \\%
27B, PRO 6000, round 1 & 7.0 & 24.3 & 26 & 1 \\%
27B, PRO 6000, round 2 & 10.4 & 37.4 & 18 & 4 \\%
Muse-Glimmer, PRO 6000 & 16.3 & 73.9 & 58 & 8 \\%
\midrule%
\multicolumn{5}{@{}l}{\textit{Batching, H100}} \\%
27B FP8 & 17.9 & 86.2 & 68 & 4 \\%
27B FP8, DSpark & 9.8 & 21.4 & 17 & 1 \\%
27B BF16, round 1 & 19.3 & 40.3 & 40 & 3 \\%
27B BF16, round 2 & 15.2 & 58.1 & 38 & 4 \\%
Qwen3.5-4B & 9.7 & 20.2 & 19 & 2 \\%
Gemma-4-12B, EAGLE-3 & 14.2 & 55.7 & 48 & 6 \\%
\bottomrule%
\end{tabular}}
\end{minipage}
\end{table}

Table~\ref{tab:app-chain} lists the fifteen promotions behind Figure~\ref{fig:evolution}. Kernel and byte work makes up most of the chain. The ceilings that opened each round match the shift described in \S\ref{sec:a-rounds}. Computed with the oracle's $\tau$ and floors at full peak rates, before round one the 2.57$\times$ ceiling split into a $\tau$ factor of 1.83$\times$, an overhead factor of 1.01$\times$ and a kernel factor of 1.39$\times$. Before round two, the 1.95$\times$ ceiling split into a $\tau$ factor of 1.34$\times$ and a step-cost factor of 1.45$\times$, so step cost led.

\section{Trees and Two-Level Verification}

\label{app:tree-tlv}

\begin{figure}[t]
\centering
\includegraphics[width=0.6\textwidth]{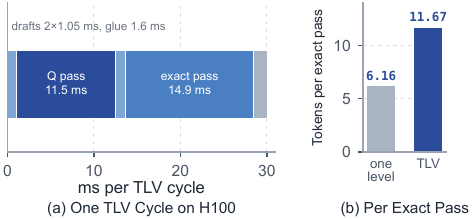}
\vskip -0.06in
\caption{Two-level verification of single-request DFlash2 on Qwen3.8-27B. (a) Time of one TLV cycle of the final system on H100. (b) Tokens accepted per exact pass in paired runs, with one level of verification and with TLV.}
\label{fig:app-tree-tlv}
\vskip -0.1in
\end{figure}

\begin{figure*}[t]
\centering
\includegraphics[width=\textwidth]{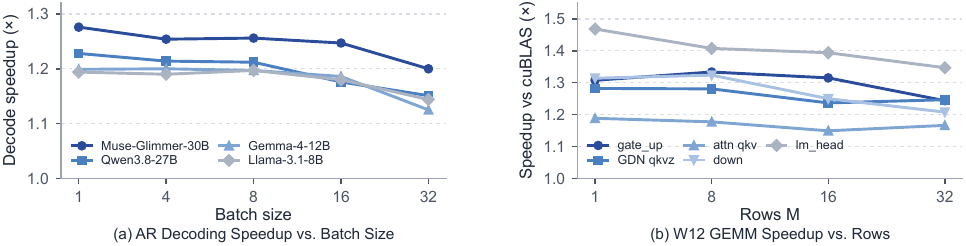}
\vskip -0.06in
\caption{W12 on H100. (a) AR decoding speedup of W12 over BF16 weights in SGLang versus batch size. (b) Speedup of the W12 GEMM over cuBLAS on Qwen3.8-27B projections versus rows, with cold L2.}
\label{fig:app-w12}
\vskip -0.1in
\end{figure*}

A TLV cycle of the final H100 system takes about 30\,ms (Figure~\ref{fig:app-tree-tlv}a). Its two drafts take 1.05\,ms each, the Q pass over the 8-bit view of W12 takes 11.5\,ms, the exact pass 14.9\,ms, and the glue between them about 1.6\,ms. In exchange, each exact pass accepts 11.67 tokens instead of 6.16 in a one-level step (Figure~\ref{fig:app-tree-tlv}b).

The Q pass can read the 8-bit view nested in the W12 records, which costs no extra memory, or a separate low-precision copy (Table~\ref{tab:app-qview}). Each view agrees with the exact pass on about 98\% of positions, and each raised speed when it was promoted. On the RTX PRO 6000, TLV with split-KV attention gained 1.147$\times$ in paired runs.

\section{Cost of the Loop}

\label{app:cost}

A round takes 7--20 hours and 20--86 GPU-hours (Table~\ref{tab:app-cost}). The thirteen completed rounds took 164 hours and 530 GPU-hours in total, created 486 code changes and promoted 54 of them. A setting reached in two rounds splits its work between them, and the second round starts from the final system of the first round.

\section{W12 Details}

\label{app:w12}

\begin{table}[t]
\centering
\begin{minipage}[t]{216.9pt}\vspace{0pt}
\caption{Weights read by the TLV Q pass: top-1 agreement with the exact pass over 104k teacher-forced positions on Qwen3.8-27B, and the paired gain when each view was promoted.}
\label{tab:app-qview}
\end{minipage}\hspace{18pt}%
\begin{minipage}[t]{221.3pt}\vspace{0pt}
\caption{AR decoding in SGLang on H100, BF16 / W12: time per output token with 1 and 32 requests, weight memory, and the KV-pool growth with W12. W12 converts every decoder linear layer and the LM head.}
\label{tab:app-w12-dec}
\end{minipage}

\vskip 0.03in
\begin{minipage}[t]{216.9pt}\vspace{0pt}
\scalebox{0.90}{%
\renewcommand{\arraystretch}{1.620}%
\footnotesize%
\setlength{\tabcolsep}{3pt}%
\begin{tabular}{@{}lccc@{}}%
\toprule%
Q view & Agreement & Deployment & Gain \\%
\midrule%
Nested 8-bit (W12) & 0.983 & H100, DFlash2 & 1.037$\times$ \\%
INT6 copy & 0.981 & H100, DSpark & 1.067$\times$ \\%
INT6/INT8 copy & 0.984 & PRO 6000, DFlash2 & 1.081$\times$ \\%
\bottomrule%
\end{tabular}}
\end{minipage}\hspace{18pt}%
\begin{minipage}[t]{221.3pt}\vspace{0pt}
\scalebox{0.90}{%
\footnotesize%
\setlength{\tabcolsep}{2.5pt}%
\begin{tabular}{@{}lcccc@{}}%
\toprule%
 & \multicolumn{2}{c}{TPOT (ms)} & Weights & KV \\%
\cmidrule(lr){2-3}%
Model & 1 req. & 32 req. & (GB) & pool \\%
\midrule%
Muse-Glimmer-30B & 20.9 / 16.4 & 21.4 / 17.8 & 60.1 / 47.7 & 2.54$\times$ \\%
Qwen3.8-27B & 19.1 / 15.6 & 23.6 / 20.5 & 51.1 / 39.8 & 2.03$\times$ \\%
Gemma-4-12B & 10.2 / 8.5 & 11.8 / 10.5 & 23.1 / 20.0 & -- \\%
Llama-3.1-8B & 6.1 / 5.1 & 6.9 / 6.0 & 16.1 / 12.7 & -- \\%
\bottomrule%
\end{tabular}}
\end{minipage}
\end{table}

W12 encodes every dense checkpoint we tried, from six model families, at 12.013--12.020 bits per weight (Table~\ref{tab:app-w12-enc}). Fewer than $1.3\times10^{-4}$ of the weights fall outside their channel's 16-binade window, so the escape list stays small.

In AR decoding, W12 keeps its gain as the batch size grows, from 1.19--1.28$\times$ with one request to 1.13--1.20$\times$ with 32 (Figure~\ref{fig:app-w12}a, Table~\ref{tab:app-w12-dec}). The freed memory goes to the KV cache, whose pool grows 2.03$\times$ on Qwen3.8-27B and 2.54$\times$ on Muse-Glimmer-30B. W12 is not tied to one serving engine. In Hugging Face Transformers with a static cache and CUDA graphs, it decodes Llama-3.1-8B 1.12$\times$ faster with one request.

The W12 GEMM beats cuBLAS on every Qwen3.8-27B projection from 1 to 32 rows (Figure~\ref{fig:app-w12}b). The LM head, the largest matrix, gains most, 1.35--1.47$\times$. With a warm L2, the speedups at up to 32 rows are 1.08--1.45$\times$. On the RTX PRO 6000, which lacks \texttt{wgmma}, the W12 GEMM runs on \texttt{mma.sync} and beats cuBLAS on all 13 projection shapes of Qwen3.8-27B and Muse-Glimmer-30B, by 1.18--1.68$\times$ up to 32 rows and by 1.02--1.68$\times$ at 48 and 64 rows. Its time stays flat in the number of rows, while cuBLAS slows down from 8 rows on.

\section{Directions the Agents Closed}
\label{app:closed}

\begin{table}[h!]
\caption{Directions closed on a measured bound, single-request DFlash2 on Qwen3.8-27B on H100.}
\label{tab:app-closed}
\vskip 0.03in
\centering
\scalebox{0.90}{%
\footnotesize%
\setlength{\tabcolsep}{2.0pt}%
\begin{tabular}{@{}llr@{}}%
\toprule%
Direction & Evidence & Result \\%
\midrule%
\multicolumn{3}{@{}l}{\textit{Weights and kernels}} \\%
11-bit weights & microbench. & 3$\times$ slower decode \\%
Persistent cross-GEMM W12 & microbench. & 196 vs 172\,$\mu$s/layer \\%
One-kernel GDN layer & microbench. & 32.3 vs 17.3\,$\mu$s/layer \\%
\multicolumn{3}{@{}l}{\textit{Width and verification}} \\%
48-node trees on W12 & timed & $-$4.3\% \\%
Three-level verification & replay & $-$2 to $-$12\% \\%
7-bit nested Q view & replay & $-$3.8\% \\%
6-bit nested Q view & replay & $-$17.5\% \\%
\bottomrule%
\end{tabular}}
\vskip -0.1in
\end{table}

The agents closed most directions on a measured bound (Table~\ref{tab:app-closed}). Kernel directions failed because the W12 GEMM is bound by decode and SM issue rate. Trees wider than 32 nodes cost more than the lattice returned: on the W12 system, each node above 32 costs about 0.17\,ms per step, against a value of about 0.063\,ms. For the nested Q view, each eighth of its bytes removed cost about 0.01 of top-1 agreement, twice what the saved time bought back.

\end{document}